\documentclass[12pt]{article}

\usepackage{chngcntr}

\usepackage{amsmath,amsfonts,amssymb, amsthm}
\usepackage{hyperref} % \url, input url
\hypersetup{colorlinks=true, citecolor=blue}
\usepackage{multirow} % various type of table
\usepackage{caption} % figure cation
\usepackage{kotex}
\usepackage{xcolor} % to use alert
\usepackage[sort]{natbib}
\usepackage{fullpage} % use full page
\usepackage{float} % to use "H" in table
\usepackage{chngpage} % to use "adjustwidth"
\usepackage{lscape}
\usepackage{mathtools}
\usepackage[blocks]{authblk}
\usepackage{booktabs} % to use \toprule in table
\usepackage{pbox} % to use line break in table
\usepackage{algpseudocode, algorithm}
\usepackage{enumerate}
\usepackage{bigints}
\usepackage{tikz}
\usepackage{caption} % figure cation
\usepackage{kotex}
\usepackage{float} % to use "H" in table
\usepackage{chngpage} % to use "adjustwidth"
\usepackage{authblk} % multiple affiliations

\hypersetup{colorlinks=true, citecolor=blue}

\title{Multiple Instance Neural Networks Based on Sparse Attention for Cancer Detection using T-cell Receptor Sequences}

\author[1]{Younghoon Kim}

\author[2,3]{Tao Wang}

\author[4]{Danyi Xiong}

\author[4]{Xinlei Wang}

\author[5]{Seongoh Park\footnote{To whom all correspondence should be addressed. Email: \texttt{spark6@sungshin.ac.kr}}}

\affil[1]{Department of Industrial and Management Systems Engineering, Kyung Hee University, Gyeonggi, Korea}

\affil[2]{Quantitative Biomedical Research Center, Department of Population and Data Sciences, University of Texas Southwestern Medical Center}

\affil[3]{Center for the Genetics of Host Defense, University of Texas Southwestern Medical Center}

\affil[4]{Department of Statistical Science, Southern Methodist University
}

\affil[5]{School of Mathematics, Statistics and Data Science, Sungshin Women's University, Seoul, Korea}

\begin{document}
\maketitle
%\tableofcontents

\begin{abstract} 
	\noindent 
%% Text of abstract
% \textbf{Motivation:}
    Early detection of cancers has been much explored due to its paramount importance in biomedical fields. Among different types of data used to answer this biological question, studies based on T cell receptors (TCRs) are under recent spotlight due to the growing appreciation of the roles of the host immunity system in tumor biology. However, the one-to-many correspondence between a patient and multiple TCR sequences  hinders researchers from simply adopting classical statistical/machine learning methods. There were recent attempts to model this type of data in the context of multiple instance learning (MIL). 
    Despite the novel application of MIL to cancer detection using TCR sequences and the demonstrated adequate performance in several tumor types, there is still room for improvement, especially for certain cancer types. Furthermore, explainable neural network models are not fully investigated for this application.
    % \noindent \textbf{Methods:}
    In this article, we propose multiple instance neural networks based on sparse attention (MINN-SA) to enhance the performance in cancer detection and explainability. The sparse attention structure drops out uninformative instances in each bag, achieving both interpretability and better predictive performance in combination with the skip connection.
    % \noindent \textbf{Results:}
    Our experiments show that MINN-SA yields the highest area under the ROC curve (AUC) scores on average measured across 10 different types of cancers, compared to existing MIL approaches. Moreover, we observe from the estimated attentions that MINN-SA can identify the TCRs that are specific for tumor antigens in the same T cell repertoire.
	
\end{abstract} 
\baselineskip 18pt

%\linenumbers

%% main text
\section{Introduction}
\label{sec:introduction}

%% Very gentle introduction to multiple instance learning
Multiple instance learning (MIL) is a supervised learning task that includes a special structure called a bag in each entity. In MIL, a set of instances in the same bag and their explanatory variables are observed. 
Though they share an observed bag-level response (bag label), instances may not have an individual instance-level response that one observes in traditional single-instance learning.
% However, they may or may not have individual response variables that we usually do not observe, but share an observed bag-level response variable called a (bag) label in common. 
Supervised MIL tasks can be classified by types of the label; multiple instance classification (MIC) if the label takes its value on a discrete space, and
multiple instance regression (MIR) if on a set of (non-discrete) real numbers. 
% As MIL field has originated from a discriminant analysis of particular molecules \cite{Dietterich:1996}, 
Most MIL applications are MIC, including remote sensing \citep{Wang:2008,Trabelsi:2019}, computer vision \citep{Sun:2016}, sentimental analysis \citep{Angelidis:2018}, and especially biology \citep{Bandyopadhyay:2015, Gao:2017,Xiong:2021}.
In the meanwhile, MIR has relatively scarce literature. We refer readers to \cite{Carbonneau:2018}  for more examples in MIC and \cite{Park:2020} in MIR.

%% Applications of MIL
%% recent advances in methodology and benchmarking data

Motivated by such applications in MIC, 
researchers have proposed a collection of methods to answer scientific questions arising from diverse real-life problems. Many of those are established by extending classical machine learning algorithms such as support vector machines \citep{zhang2007local,andrews2003support,gartner2002multi,ray2005supervised}, K-nearest neighbors \citep{Carbonneau:2018, wang2000solving}, and gradient boosting \citep{babenko2008simultaneous}. The key concepts underlying their extensions are reviewed by \cite{Amores:2013} where the author proposed a new taxonomy based on how a method handles a set of instances. More details about the algorithms can be found in  \cite{Xiong:2021}.

% As a pioneering work that reviews classification methods in MIL, \cite{Amores:2013} classify MIC approaches based on a new taxonomy and compare them through numerical experiments. Another comparative study from \cite{Carbonneau:2018} has been recently published where they share Matlab codes of numerous methods in their Github page built upon the earlier work (\cite{Tax:2016}). 
% However, a bunch of methods/algorithms have emerged together
% In  \cite{Xiong:2021} a comparative study was conducted 

In contrast to the diversity found in methodology and application, as pointed out in \cite{Carbonneau:2018}, benchmark MIL datasets have been still limited to predicting binding sites of molecules, image/text classification in literature. For instances, \cite{Amores:2013,Carbonneau:2018} have conducted a comparative study using typical real data examples so-called Musk \citep{Dietterich:1996}, Text \citep{Andrews:2003}, Speaker \citep{Sanderson:2009}, Corel \citep{Chen:2006}, Birds \citep{Briggs:2012}, Letters \citep{Frey:1991}, and Tiger/Elephant/Fox (TEF) \citep{Andrews:2003} %HEPMASS (\citep{Baldi:2016}) 
, which other MIL researches have also used  \citep{Kim:2014,Cheung:2006,Raykar:2008,Bergeron:2012,Cheplygina:2015,Ilse:2018,Asif:2019}.
%% TCR sequencing data is indeed MIL!
% % importance of cancer detection problem (using TCR seqeucing data)
In the light of it, there were recent attempts to bring brand-new data under the MIL framework \citep{Ostmeyer:2019,Xiong:2021}: T cell receptor sequencing data for cancer detection.

The problem to distinguish normal and cancerous tissues/patients has attracted much attention due to its great significance in biomedical fields for cancer prognosis. To address this biological problem, 
% not only methods have evolved in statistics and machine learning fields, but the forms of data are also diversified. For example,
past works have used medical images \citep{Saba:2020,Yan:2021}, gene expressions \citep{Lu:2003,Li:2017,Verda:2019,Mostavi:2020}, and single nucleotide polymorphisms (SNPs) \citep{Hajiloo:2013, Batnyam:2013,Boutorh:2015}, etc. As the basis for forming predictive models, the recent appreciation of the roles of the host immunity system in tumor biology has motivated researchers to study T cell receptors (TCRs) \citep{Beshnova:2020, GEE:2018, Lu:2021, Xiong:2021,Ostmeyer:2019}. 
\cite{Xiong:2021} predicted the tumor status of patients using TCR sequences. As multiple TCR sequences (instances) are observed together in different T cells in the same patient (tumor or normal), the observations naturally fall into the category of MIL. The authors conducted a benchmarking study to compare MIC algorithms, but they did not treat deep neural networks in depth, thus leaving the performance of neural network models under-explored. Though \cite{Wang:2018_revisiting}'s models are included in comparison, they are not complex enough to learn successfully the underlying structure of the TCR data, thus showing unsatisfactory performance. 
\cite{Ostmeyer:2019} investigated the multiple instance learning task that distinguishes T-cell repertoires between
tumor and healthy tissues, but they assumed the standard MIL assumption which does not consider relative importance of instances. \cite{Beshnova:2020} proposed a deep learning model called DeepCat that utilizes tumor-specific or non-cancer TCRs, but DeepCat ignores the bag-structure as well as possibly different contributions of TCRs.
Another deep learning model, named DeepLION, is proposed by \cite{Xu:2022}, but DeepLION cannot completely remove unimportant instances in explaining the bag labels.

%% Contribution of our work
Towards bridging this gap, we introduce a novel neural network model, titled MINN-SA (Multiple Instance Neural Network based on Sparse Attentions), for cancer detection based on TCR sequences. The salient part of the proposal is the sparse attention structure that flexibly drops out uninformative instances, thus rendering model interpretation more achievable. Recent works from \cite{widrich2020modern,tourniaire2021attention, rymarczyk2021kernel, lu2021smile} also employed an attention structure in their neural networks. However, their attention scores are dense, meaning none of them is exactly 0, so that irrelevant information for bag classification could be involved in extracted features. Moreover, the sparsity pattern considered in \cite{lu2021smile} is based on a simple heuristic that keeps top-$N$ instances with the largest scores, which lacks of optimality and stability in results. In contrast, equipped with the sparsemax function by \cite{Martins:2016}, MINN-SA adaptively discovers the pattern of sparsity in attention scores. This flexibility is also beneficial to predictive performance, which can be further enhanced by adding the skip connection by \cite{He:2016}. With this state-of-the-art architecture, we achieve the highest overall AUC scores both in balanced and imbalanced datasets in the cancer detection problem. 
Our main contributions are summarized as follows:
\begin{itemize}
    \item We propose a sparse attention-based neural network that drops out uninformative instance per bag, achieving model interpretability.
    
    \item MINN-SA outperforms comparative methods in cancer detection based on TCR sequences, achieving the highest AUC scores both in balanced and imbalanced datasets.

\end{itemize}

% The main contributions of this study can be summarized as follows: 
% \begin{enumerate}[(1)]
%     \item We propose a sparse attention-based neural network for MIL. To the best of our knowledge, this has not been studied in previous work. Our method aims to determine important receptors of the T cell for cancer classification.
%     \item We use sparsemax in the attention module to force redundant instances' attention values equal to zero. The output can be efficiently calculated with the projection of dense attention scores.
%     \item To demonstrate the usefulness of the proposed method, we examined the performance of the proposed method compared with previous approaches. The results demonstrate that our proposed method outperforms the alternatives regarding predictive accuracy, interpretability, and computational efficiency.
% \end{enumerate}

The remainder of this paper is organized as follows. Section 2 gives the details of our method. Section 3 presents the experimental setup and results of TCR dataset. Finally, we end in Section 4 with a summary and discussion.

\section{Methods}
\label{sec:methods}
\subsection{Multiple instance learning for cancer detection using TCR sequences}
In MIL, an observational unit is a bag (a sample). In bag $i$ ($i=1,\ldots, n$), each of multiple instances is characterized by a vector $x_{ij}$ of $p$ features in $\mathbb{R}^p$, $j=1,\ldots, m_i$, and a single label $y_i$ is tagged on it. The goal of MIL is to estimate a function $f$ that predicts a bag-level label from a set of instances. Note that this function takes a set of instances as an input, so it should be adaptive to different number of instances for each bag.

In our application, tissue samples are collected either from normal or cancer patients and a set of TCR sequences are identified in each sample by using next generation sequencing technologies (\cite{Xiong:2021,Zhang:2021}). The main task is to determine whether a tissue is cancerous or not based on its TCR sequences. Here, we treat the tissue type as a bag-level label and the set of TCR sequences as multiple instances, all of which are contained together in a patient, or a bag. Under this context, we focus on a binary MIC task and thus restrict a bag label in $\{0,1\}$; for example, 0 (a negative bag) is non-cancer and 1 (a positive bag) is cancer. To associate a series of unlabeled instances to a bag label, we adopt the primary instance assumption. In other words, it is assumed that a portion of instances, or primary instances, can explain the label while the remaining instances, or non-primary instances, are irrelevant to it.
In our contexts, those selected TCRs represent specialized T cells that the human immune system develops against the tumor cells. More specifically, the TCRs recognize the tumor-associated antigens (\cite{lee1999characterization,LEWIS:2003}) or tumor neoantigens (\cite{Gubin_2015,Stevanovi_2017,Lu:2020sciimmunol}) presented on the surface of the tumor cells, which are markers of the tumor cells and distinguish them from normal epithelial cells.

We utilize the sparse attention in the multiple instance neural networks to detect such meaningful TCRs. The proposed layer selectively reflects instances' information to an extracted feature vector for final classification. The sparsity enhances the classifier's performance and the explainability of classification results. Moreover, the proposed method is computationally efficient. The details of the proposed method are presented in the following subsections.

\subsection{Numeric embeddings of TCR sequences}\label{sec:data_generation}

We describe the process of numeric embedding of TCR sequences carried out in \cite{Xiong:2021}. TCR sequence is a text string comprising a series of amino acids, which is actually a text string. According to \cite{Atchley:2005}, each amino acid can be converted to five Atchley (latent) factors that sufficiently represent the attributes of the amino acid. This conversion of a set of TCR sequences returns a Atchley matrix, which is inserted into the TCR encoding algorithm \citep{Zhang:2021, Lu:2021}. The key part of the algorithm is a stacked auto-encoder that takes Atchley matrices and returns a set of vectors of fixed length  (30 dimensions) determined by the number of neurons in the bottleneck layer. The encoded numeric representation facilitates the usage of TCR sequence data. Most MIL algorithms are only compatible with the numeric type of data, especially for those methods calculating a distance between instances (or bags). We refer to \cite{Xiong:2021} for more information about the data and processing details. In particular, we do not claim any original contribution to the data.

\subsection{Neural networks based on sparse attention}\label{sec:model}

\begin{figure}[H]
	\centering
	\includegraphics[width=1\linewidth]{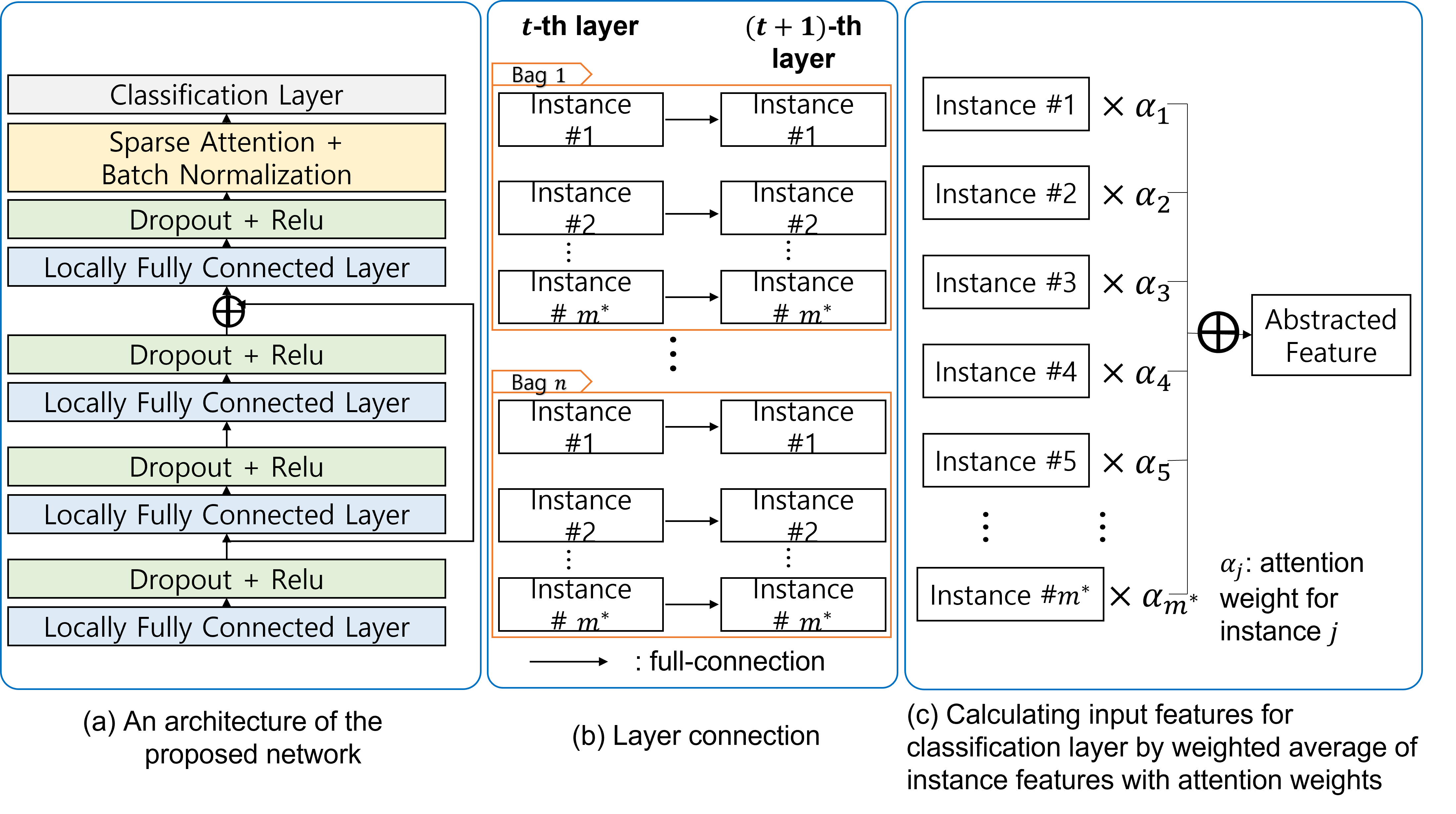}
	\caption{Description of the proposed neural network.}
	\label{fig:architecture}
\end{figure}

We propose a neural network based on sparse attention to solve multiple instance classification problems. The overall structure of our model is illustrated in Figure \ref{fig:architecture} (a), and we give details of each component in our neural network below. 

The input size is fixed by $m^* \times p$ for each of $n$ bags where $m^*$ is the hyperparameter that decides the number of instances to be included in modeling. If a bag has fewer instances than $m^*$, then an empty part is padded by zeros. Some of bags have larger instance size than $m^*$. In the case, we determine the first $m^*$ instances as our bag instances.  Also, we keep using this masking information across the whole process. For ease of handling, one can easily set $m^*$ by the largest bag size in data.  In Section \ref{sec:results}, we conduct a sensitivity analysis for choice of the size $m^*$.

Each layer consists of $m^* \times p$ neurons, but they are not fully connected to the activation functions in the previous layer as in the usual fully-connected layer. Instead, we build the full connections between neurons within each instance, but not across different instances (see Figure \ref{fig:architecture} (b)). The weights for a fully connected layer are shared across the instances to handle the variable number of instances in each bag. Hence, the two consecutive layers are connected by a $p\times p$ weight matrix (or $(p+1)\times (p+1)$ if a bias term is included). The network deals with the non-linearity in data with the rectified linear unit (ReLU) \citep{Nair:2010}. Note that we used fully connected layers rather than convolution layers because the input features are not locally correlated between adjacent features.

We employed the dropout \citep{Srivastava:2014} and skip connection \citep{He:2016} to enhance the predictive performance. The dropout layer is attached after each locally fully-connected layer appears. It randomly forces the output variables to be zero with probability 0.3 while training the network. It is well known that this layer prevents the complex neural network from overfitting on training data. The residual learning framework eases the training of networks with stacked layers. The skip connections in the framework explicitly let the layers fit a residual mapping instead of hoping stacked layers to directly fit a desired underlying mapping denoted by $\mathcal{H}(x)$. We let the stacked nonlinear layers fit another mapping of $\mathcal{F}(x):=\mathcal{H}(x)-x$. The bypassing path for a gradient mitigates the vanishing gradient issue in neural networks. The empirical studies show that it is easier to optimize the residual mapping than to optimize the original. Thus, we employ the skip connection to enhance the performance of the locally fully-connected neural network with multi-layers. The effect of residual connection is demonstrated in Section \ref{sec:ablation}.

The attention layer combines the attention weights $\{\alpha_j\}_{j=1}^{m^*}$ returned by the sparsemax function \citep{Martins:2016} with the feature matrix $Z\in \mathbb{R}^{m^* \times p}$ obtained from the last network layer, ending up with a weighted feature vector $\tilde{z}=\sum_{j=1}^{m^*} \alpha_j z_j$ where $z_j$ is the $j$-th row of $Z$ and $\tilde{z}\in \mathbb{R}^{1 \times p}$ (see Figure \ref{fig:architecture} (c)). The commonly used softmax function does not pursue exact zeros in the output, but the sparsemax function permits zeros in it. Let $\Delta^{K-1} := \left\{ p\in \mathbb{R}^K|1^Tp=1, p\ge 0 \right\}$ be the $(K-1)$-simplex. The sparsemax is a function mapping vectors in $\mathbb{R}^K$ to probability distributions in $\Delta^{K-1}$: 

\begin{equation} \label{eu_eqn}
\textrm{sparsemax}(z):=\underset{p\in\Delta^{K-1}}{\textrm{argmin}} \left\| p-z \right\|^2.
\end{equation}
    
In our context, the sparsemax function takes attention scores and encourages some of them to be zero if they do not exceed some thresholding value. As shown in \cite{Martins:2016}, the threshold is adaptively determined from the scores, not manually by users like hard/soft-thresholding functions.

For each bag, two scores, which correspond to classes of tumor and normal tissues, are converted to probabilities, which is easier to interpret. The transition is done by the sigmoid function. The input of classification layer is batch-normalized feature vector $\tilde{z}^{*}\in \mathbb{R}^{1 \times p}$ and the output is a scalar. The aggregated feature vector $\tilde{z}$ may have different means and variances across components, which could hamper the network from learning data stably. Thus, we apply the batch normalization \citep{ioffe2015batch} to overcome this difficulty.

We should mention the difference between the proposed attention structure and the others from the previous works. Derived from \cite{widrich2020modern,tourniaire2021attention, rymarczyk2021kernel, lu2021smile}, the attention scores are strictly larger than zero. Thus, it is challenging to discard unimportant or non-primary instances that have little to do with bag classification. In contrast, MINN-SA allows the attention scores to be sparse or have many zeros, meaning that strictly positive weights are only given to the primary instances responsible to bag classification. This is an attractive instance selection because it transparently shows which instances are chosen in model training and thus facilitates one to interpret classification results only depending on the selected instances.
Moreover, our selection is more advanced than the elementary top-$N$ rule (\cite{lu2021smile}) and free from tuning parameters often required in thresholding operators.

\subsection{Computation}
We implement the method with a deep learning framework, PyTorch 1.8.1 \citep{paszke2019pytorch} with CUDA 11.1 \citep{garland2008parallel}. The computation system consists of Intel i9-10900 CPU, 32GB RAM, and RTX 3090 GPU. It takes 0.01 seconds for each epoch in the training stage. We stop the training after 100 epochs and determine the best performing model during the training procedure. Therefore, the whole training process approximately takes 1 second in our computation system setting. Referring to the computational time results from \cite{Xiong:2021}, the proposed method is more efficient than other multiple instances learning methods in computation.

\section{Results}
\label{sec:results}

In this section, we showcase our novel model in distinguishing tumor and normal tissue samples from different types of cancers. Different existing methods are compared against our model in terms of predictive accuracy. Moreover, we provide the instance selection result derived from the estimated attention weights. Lastly, we conduct an ablation study to examine contributions of individual components to our model.

The real datasets we analyze are from The Cancer Genome Atlas (TCGA) database, in whole generated by the TCGA Research Network: \url{https://www.cancer.gov/tcga}.
Normal and healthy tissues are collected for 10 types of cancers listed in Table \ref{tab:TCGA_label}. As mentioned in Section \ref{sec:data_generation}, these samples are processed through the next generation sequencing, TCR reconstruction techniques, and TCR encoding algorithms so that the genomic data from the donors are converted in numeric vectors.

\subsection{Setting}
% % bag distribution
To figure out how they behave in different scenarios, we test the comparative models under two scenarios: (1) the balanced case and (2) the imbalanced case. The former has an equal number of positive (tumor) and negative (normal) bags, while the latter sets about 10\% of bags to be positive. 
The balanced data is commonly used and often preferred in machine learning literature. The other one is to capture characteristics of large population cancer screening where few patients have tumors \citep{Lin:2012,Fotouhi:2019,Xiong:2021}.
To create a dataset for each cancer type, we subsample normal and tumor tissues to keep the aimed proportion of positive (tumor) bags. The sample size of each dataset is tabulated in Table \ref{tab:TCGA_label}.
\begin{table}[H]
\centering
% 		\begin{adjustwidth}{-2.0cm}{}
	\begin{tabular}{|c|ccccc|}
\hline
\bf Cancer &\bf BRCA &\bf DLBC &\bf ESCA &\bf KIRC &\bf LUAD\\
\hline
Balance & 404 & 90 & 332 & 404 & 404 \\
Imbalance & 225 & 225 & 225 & 225 & 225 \\
\hline
\hline
\bf Cancer & \bf LUSC &\bf OV &\bf SKCM &\bf STAD &\bf THYM\\
\hline
Balance & 404 & 404 & 404 & 404 & 216\\

Imbalance & 225 & 225 & 225 & 225 & 225\\
\hline
\end{tabular}
% \end{adjustwidth}
	\caption{The sample size for each cancer type. Half of the samples are tumor tissue samples for the balanced case, while about a tenth of them are for the imbalanced case.}
	\label{tab:TCGA_label}
\end{table}
\noindent
For model training and validation, we conduct 10-fold cross validation (CV) to split training and testing datasets. On the testing dataset, the Area Under the Curve (AUC) of each method is calculated based on the Receiver Operating Characteristic (ROC) curve. We follow the same experimental design in the preceding work \citep{Xiong:2021} for fair comparison.

\subsection{Benchmarking on cancer detection}
% % method
To benchmark the proposed model, 18 MIC methods are considered, which are listed in Table \ref{tab:method}. We refer to Section 3 of \cite{Xiong:2021} for a detailed exposition about these methods.

\begin{table}[H]
	\begin{adjustwidth}{-0.7cm}{}
	\small
\begin{tabular}{|ccc|}
	\hline
	\multicolumn{3}{|c|}{\bf 18 comparative methods} \\
	\hline\hline
	ADeep \citep{Ilse:2018} & 
	BoW \citep{Amores:2013} & 
	CCE \citep{zhou2007solving}\\\hline 
	CkNN \citep{wang2000solving}&
	EMD-SVM \citep{zhang2007local} &
    EMDD \citep{zhang2002dd} \\\hline 
	SI-kNN \citep{Carbonneau:2018} &  MI-SVM \citep{andrews2003support} &
	 miGraph \citep{zhou2009multi} \\\hline 
	 MILBoost \citep{babenko2008simultaneous} &
	MILES  \citep{chen2006miles} &
	MInD \citep{cheplygina2015multiple} \\\hline
	mi-Net \citep{wang2018revisiting} &  MI-Net \citep{wang2018revisiting}& 
	mi-SVM \citep{andrews2003support} \\\hline
	NSK-SVM \citep{gartner2002multi} & 
	SI-SVM \citep{ray2005supervised} & MINN-SA \\
	\hline
\end{tabular}
	\end{adjustwidth}
	\caption{Abbreviations of 18 comparative methods including the proposed method ``MINN-SA'' and their original references.}
	\label{tab:method}
\end{table}

\begin{figure}[H]
	\centering
	\includegraphics[width=1\linewidth]{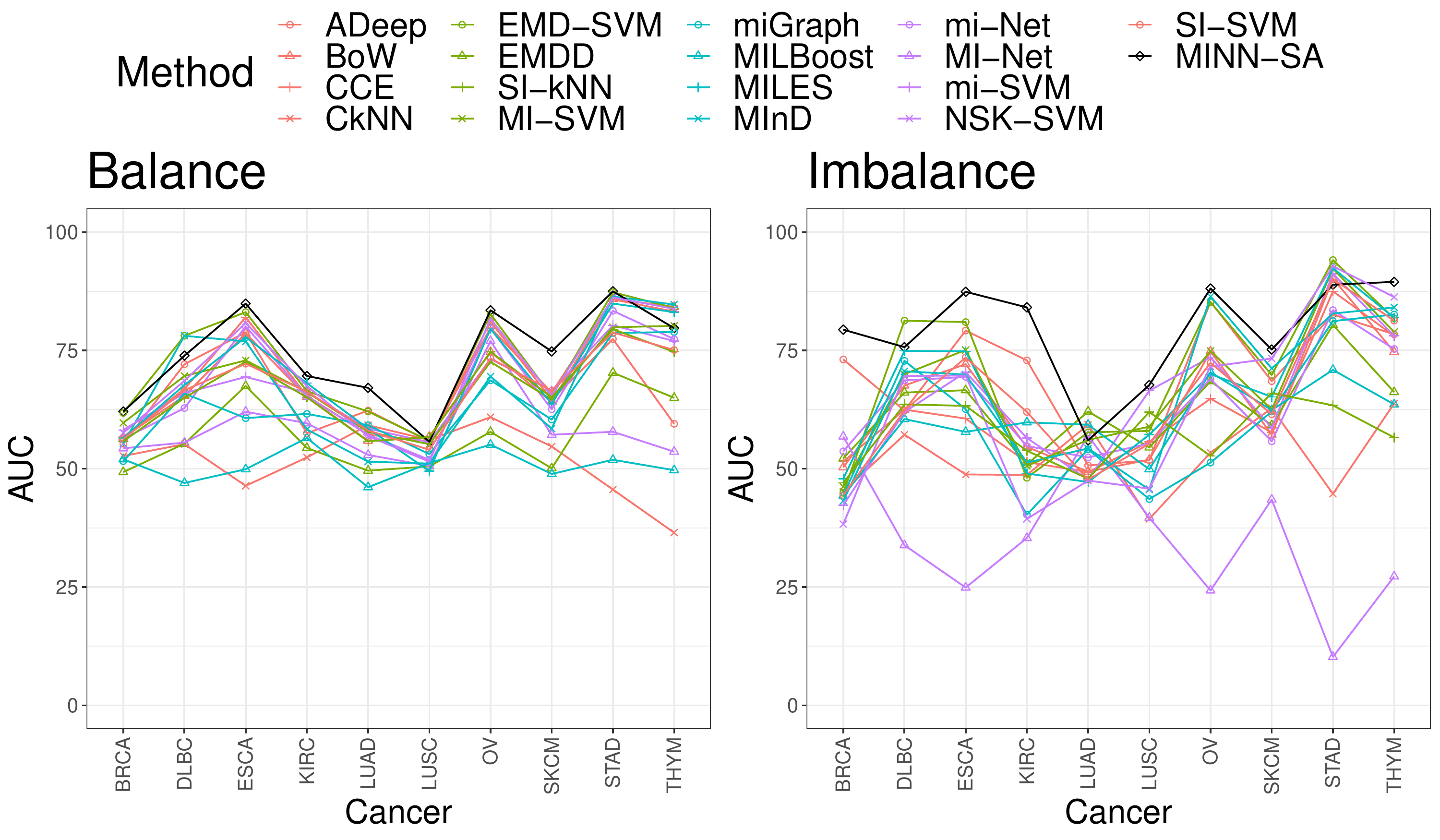}
	\caption{The AUC of all models in comparison across different cancer types. Each point is the average AUC values over 10-fold CV.}
	\label{fig:AUC_by_cancer}
\end{figure}

Figure \ref{fig:AUC_by_cancer} shows average AUC values by methods for the ten cancer types. Remarkably, MINN-SA dominates all methods in most of cancer types. Out of 10, MINN-SA wins in 7 types (BRCA, ESCA, KIRC, LUAD, OV, SKCM, STAD) for the balanced case and in different 7 types (BRCA, ESCA, KIRC, LUSC, OV, SKCM, THYM) for the imbalanced case. KIRC, LUAD, SKCM, LUSC are the four most immunogenic cancer types \citep{Wang:2018}, meaning they have a lot of T cell infiltrations. It makes sense these cancer types are among the ones  \citep{Wang:2018} for which our model performs the best, which investigates TCRs of T cells for classification. Generally, when the class distribution is balanced, each model shows more stable performance \citep{Wang:2018}. The gap between MINN-SA and the second best method is considerably big in BRCA and KIRC datasets for the imbalanced case.

\begin{figure}[H]
	\centering
	\includegraphics[width=1\linewidth]{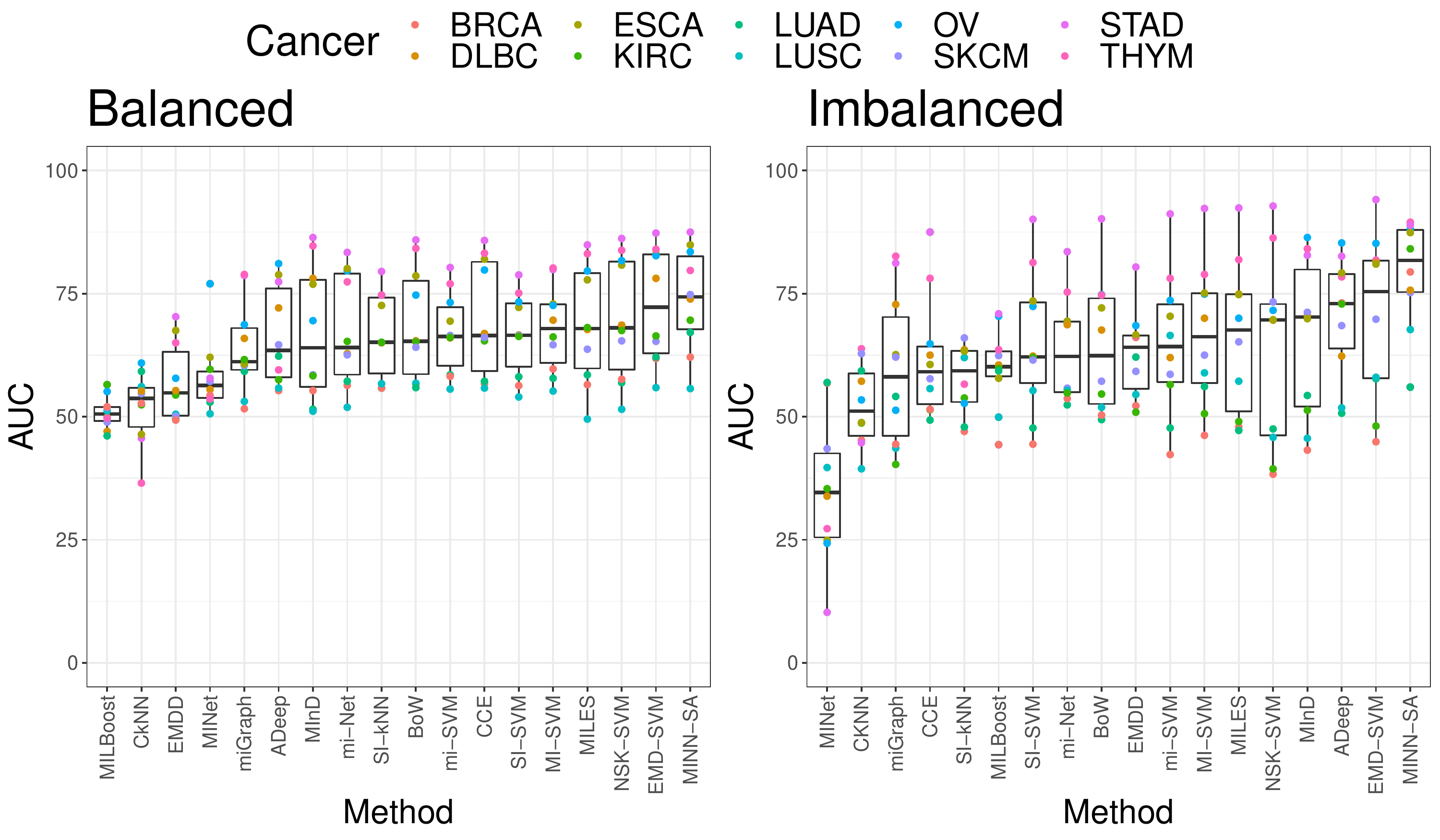}
	\caption{Boxplots of the AUC values across different methods in comparison. 
	They are ordered by median AUC.
	Each point is the average AUC values over 10-fold CV and different colors are used to show types of cancer.}
	\label{fig:AUC_by_method}
\end{figure}

Figure \ref{fig:AUC_by_method} shows the boxplots of all methods in an ascending order of median AUC. MINN-SA tops in both balanced imbalanced cases, with medians 74.40 and 81.80, respectively, followed by EMD-SVM with 72.20 and 75.40. To demonstrate the superiority of the proposed method over comparative methods, we conducted Wilcoxon signed-rank test on the best and second-best methods with rank statistics. Our proposed method achieves the best performance in terms of average rank over cancers. The average ranks of the proposed method are 2.3 and 2.2 in balanced and imbalanced cases, respectively. The second best method is EMD-SVM, whose average ranks are 2.7 and 5.2. In the balanced case, the p-value is 0.0372, and the p-value for the imbalanced case is 0.0043. Our proposed method outperforms other approaches in both cases with statistical significance.

Moreover, the proposed MINN-SA enjoys interpretable results from selected instances, which EMD-SVM does not afford. Also, MINN-SA is not very sensitive to class imbalance. Contrary to it, most MIL methods have degraded performance for the imbalanced case, calling for further modifications; for example, data generation or modifying the loss function of the classification model \citep{Huang:2016,Sundin:2019,Yang:2020}.

\subsection{Attention of instances}
\begin{figure}[H]
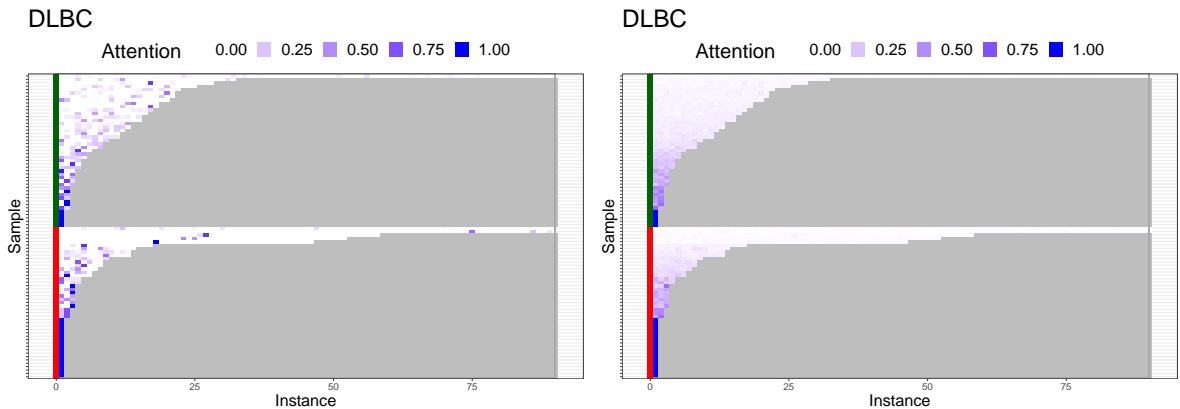

	\centering
%	\begin{minipage}{0.5\linewidth}
		\includegraphics[page=2,width=0.47\linewidth]{./figures/220715_Atten_heatmap_label_90_balanced_1.pdf}
		\includegraphics[page=2,width=0.47\linewidth]{./figures/220715_Atten_heatmap_label_90_balanced_softmax_1.pdf}
%	\end{minipage}
	\caption{Heatmap of estimated attention weights via the sparsemax (left) and softmax (right) function. Samples are reordered by the number of instances and the masking area (no instances) is colored in gray.
	Red and green labels indicate tumor and normal samples, respectively. Here, we show the balanced case for DLBC cancer data.}
	\label{fig:attention}
\end{figure}

In Figure \ref{fig:attention}, attention weights of instances are displayed in heatmap. The heatmap is given in a $n \times m^*$ matrix form where the weights are colored in blue-white spectrum and the masking area (no instances) in gray. The visual inspection demonstrates that the attention weights estimated by MINN-SA are sparser than those by the softmax-based method. For the case based on the softmax function, all instances have strictly positive weights, which is depicted by smooth patterns in the heatmap. Consequently, the dense weights makes the aggregated feature vector from the attention layer depend on redundant information for classification. On the other hand, MINN-SA forces the attention weights of insignificant instances to be exactly zero, which makes  decisions of MINN-SA independent of them. 
In our data, the selected instances are likely the TCRs that are specific for tumor antigens, such as tumor neoantigens or tumor associated antigens, presented on the surface of the tumor cells. 

% wilcox rank test balanced 0.0372 imbalance 0.0043

% This instance selection is concordant with the concept of primary instances in MIL literature \citep{Ray:2001, Park:2020,Subramanian:2016}. 

\subsection{Extracted Features}

\begin{figure}[H]
	\centering
%	\begin{minipage}{0.5\linewidth}
		\includegraphics[page=1,width=1\linewidth]{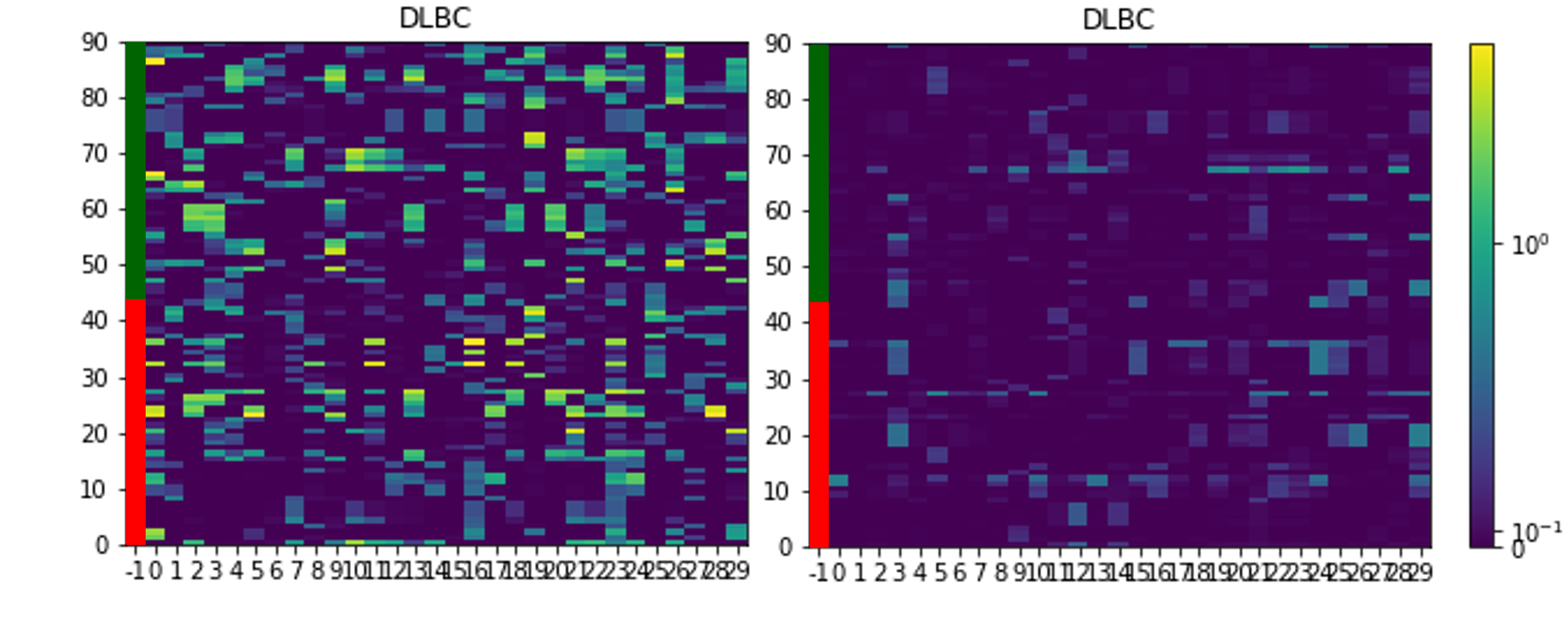}
		
%	\end{minipage}
	\caption{Heatmap of extracted features via the sparsemax (left) and softmax (right) function. Values are min-max normalized and transformed in log-scale. Red and green labels indicate tumor and normal samples, respectively. Here, we show the balanced case for DLBC cancer data.}
	\label{fig:extracted}
\end{figure}

Figure \ref{fig:extracted} shows the extracted features before the last classification layer. The heatmap is given in a $n \times p$ matrix form colored in a blue-yellow spectrum. 
It can be seen that the extracted features using the sparsemax function (left) are more activated than using the softmax function (right). Hence, the difference between the features in each observation of the sparsemax case is more distinct than the softmax case. The results demonstrate that the extracted features using the sparsemax function are more informative to characterize the characteristics of each sample. We believe that the sparsity in the proposed attention structure distinguishes the instances responsible for the bag classification so that the aggregated features can accurately discriminate bags in the classification layer.

\subsection{Ablation study}\label{sec:ablation}

Firstly, we perform an ablation study to measure contributions of the two components to our neural network: (1) the skip connection and (2) the sparsemax function. Thus, we set a baseline model, denoted by ``FC'' (short for ``fully-connected''), by removing the two components from MINN-SA, and we add each component one after another to ``FC''. This leads to the four comparative models shown in Table \ref{tab:ablation_model}. Note that ``FC'' is the model used in \cite{Ilse:2018}, and ``Proposed'' is MINN-SA.

\begin{table}[H]
	\centering
	\begin{tabular}{|c|c|c|c|c|}
		\hline
		 &\bf FC &\bf Skip &\bf Sparse &\bf Proposed \\
		\hline
        \hline
		Skip connection & $\times$ & $\checkmark$ & $\times$ & $\checkmark$ \\
		Sparsemax layer & $\times$ & $\times$ & $\checkmark$ & $\checkmark$ \\
		\hline
		Balance & 66.44 & 69.61 & 70.33 & 73.87 \\
		Imbalance & 70.47 & 73.79 & 75.61 & 79.19 \\
		\hline
	\end{tabular}
\caption{Comparison of the four models in the ablation study. Structural difference is checked in the second and third rows where ``X'' means absence of such structure and ``$\checkmark$'' means presence of it.
Their average AUC values are summarized in the last two rows.}
\label{tab:ablation_model}
\end{table}

\begin{figure}[H]
	\centering
	%	\begin{minipage}{0.5\linewidth}
	\includegraphics[page=1,width=1\linewidth]{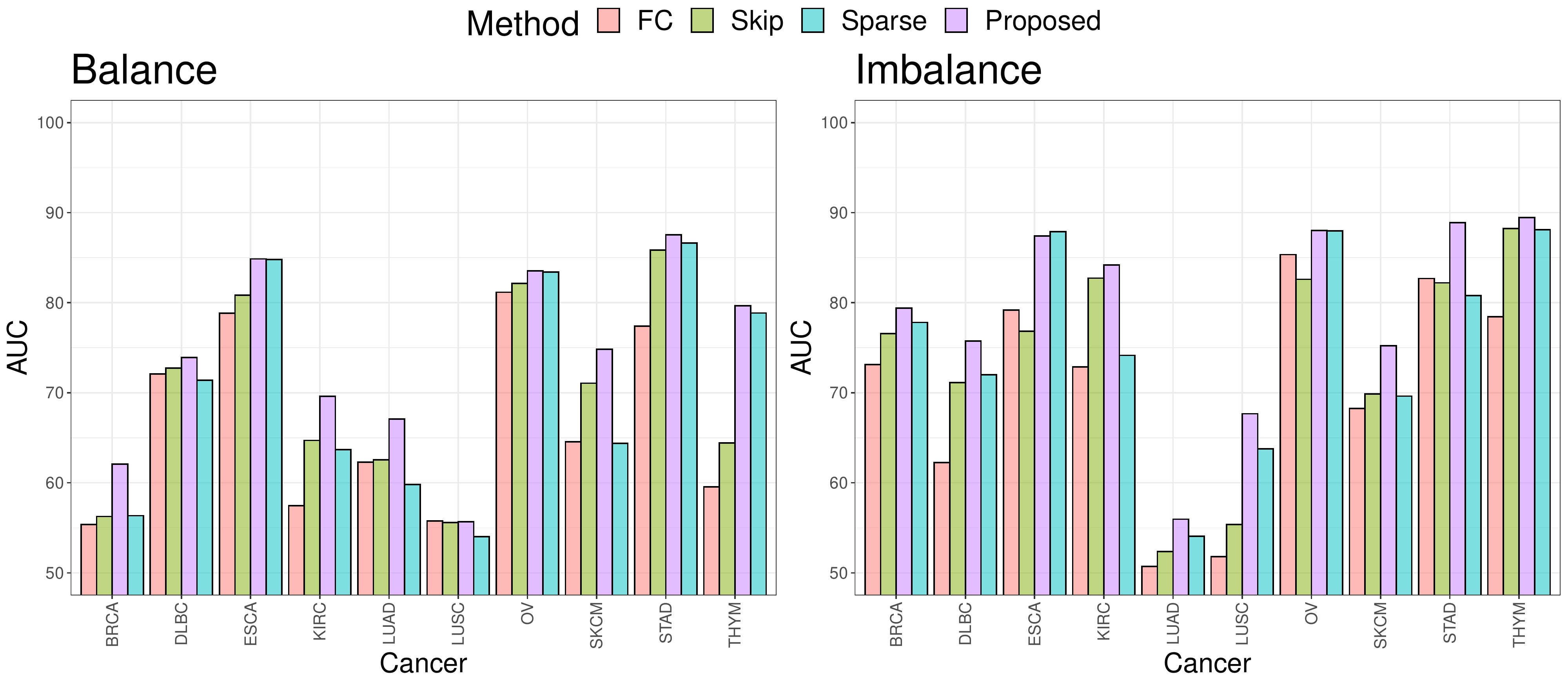}
	%	\end{minipage}
	\caption{The AUC of four models across different cancer types. Each bar denotes the average AUC values over 10-fold CV.}
	\label{fig:ablation_component}
\end{figure}

Figure \ref{fig:ablation_component} shows AUC values of the four models for different cancer types. ``Proposed'' outperforms the others in most cases; otherwise it makes a close second. The superiority of ``Proposed'' is less distinct in the imbalanced case, but it always takes the first or second place, and thus achieves the highest AUC value 79.19 in average followed by 75.61 (see Table \ref{tab:ablation_model}). These results lend strong support to the proposed method against the existing multiple instance neural network model by \cite{Ilse:2018}, a fully-connected neural network based on the softmax function without the skip connection. According to the known phenomenon of immunodominance in the field of immunology \citep{Akram:2012, Yewdell:1999}, not all instances own necessary information and thus such instances would be better removed for classification by the sparse attention structure. The skip connection also proves valuable for this specific application.  Interestingly, the performance is the best when both skip connection and sparsemax function are utilized together. This phenomenon aligns with the previous study \citep{he2019bag} that shows combining regularization tricks can achieve the best classification performance.

\begin{figure}[H]
	\centering
	%	\begin{minipage}{0.5\linewidth}
	\includegraphics[page=1,width=1\linewidth]{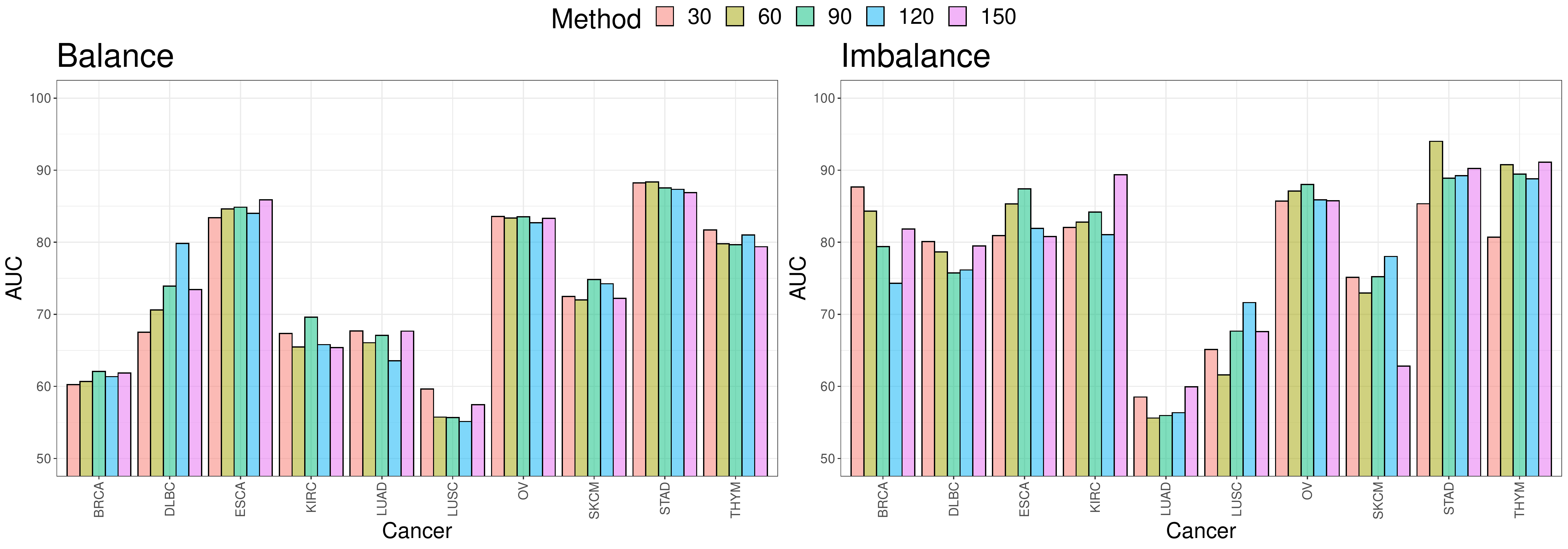}
	%	\end{minipage}
	\caption{The AUC of the proposed models having different capacity (bag size) of instances across different cancer types.  Each bar denotes the average AUC values over 10-fold CV.}
	\label{fig:ablation_length}
\end{figure}

%\subsection{Sensitivity analysis}
In the following ablation study, we assess the sensitivity to the maximum number of instances per bag. We have tried different maximum numbers by $m^*=30, 60, \ldots, 150$ and their AUC values are reported in Figure \ref{fig:ablation_length}. 

\begin{figure}[H]
	\centering
	%	\begin{minipage}{0.5\linewidth}
	\includegraphics[page=1,width=1\linewidth]{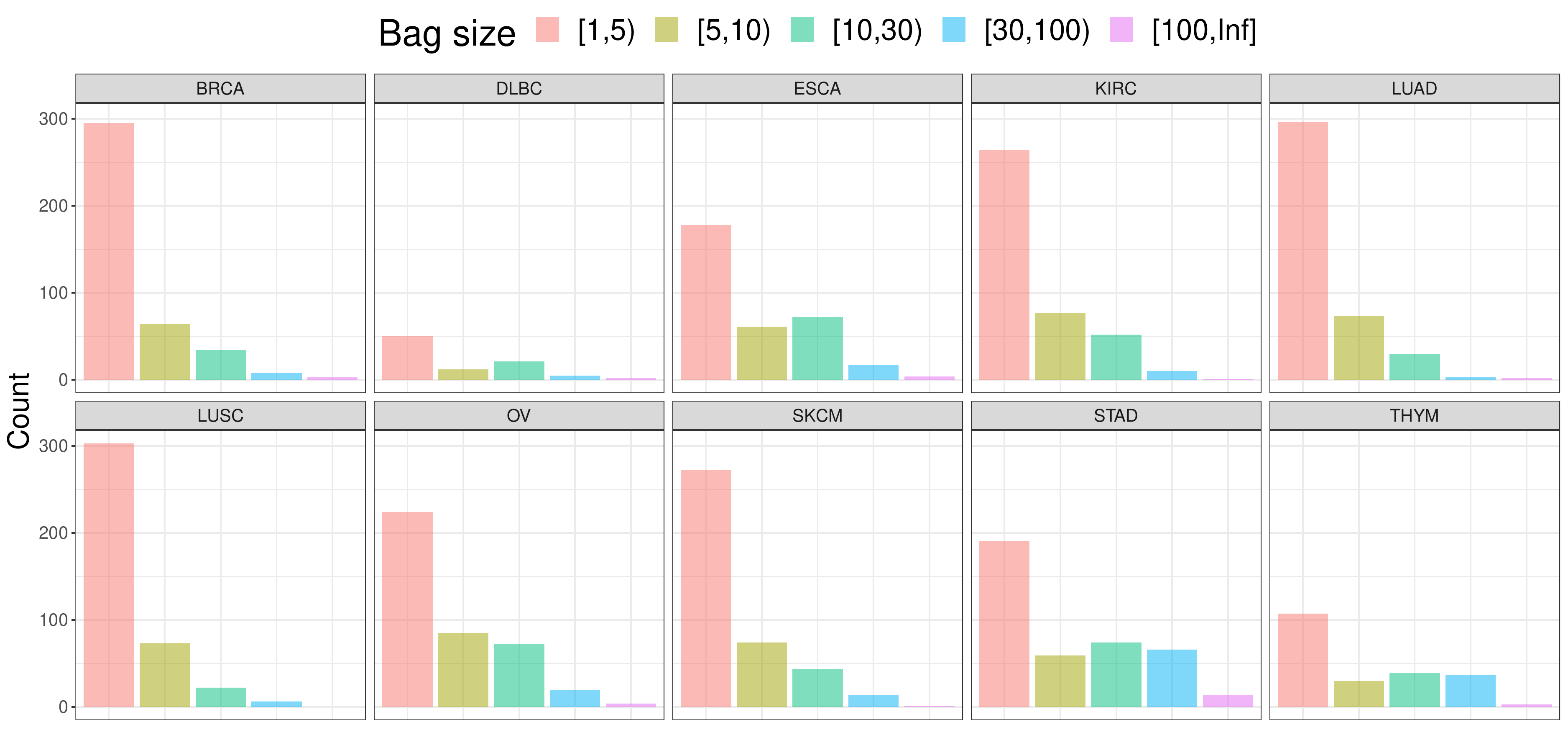}
	%	\end{minipage}
	\caption{The distribution of bag sizes in the balanced dataset. Most of bags have less than 5 instances, expressing heavy-tailed features.}
	\label{fig:bag_size}
\end{figure}

The results show that there is no significant difference in classification performance according to the hyperparmeter $m^*$. This implies that the absolute amount of data increases as $m^*$ increases, but useful information learned for classification could be limited. However, as shown in Figure \ref{fig:bag_size}, bags with more than 30 instances belong to the minority of cases, calling for caution about generalization of this result to other applications. Hence, one should take characteristics of data in hand into account to decide which range of $m^*$ would be explored. The bag size $m^*$ is a hyperparameter for the machine learning model and can be optimized by validation procedures such as $k$-fold cross-validation.

\section{Discussion}
\label{sec:discussion}

This paper shows that MINN-SA has improved performance on most cancer types compared to the existing methods in predicting tumor vs. normal tissue samples using TCRs. The results imply that a deep neural network is suitable for multiple instance learning. In contrast, traditional statistical approaches may not work that well, and this is because the deep learning approaches can capture very complicated bag-instance structures. The flexible structures of deep neural networks reflect the bag-instance information efficiently through purely data-driven approaches. For statistical approaches, such bag-instance relationships have to be assumed and sophisticatedly specified. We believe the hand-made specifications are vulnerable to data variations such as cancer types.

Potentially, the proposed model can be also applied to various MIC problems where instances are naturally arranged in sets and having weakly annotated data. MINN-SA produces interpretable results illustrating the importance of instances and thus selecting a subset of primary instances. The applications include biology and chemistry, computer vision, document classification, web mining, reinforcement learning, speech recognition, and time series classification \citep{Carbonneau:2018}. Specifically to biology and bioinformatics, the bags consist of complex chemical or biological entities. 
% The proposed method could predict drug's bioavailability, the binding affinity of peptides to major histocompatibility complex molecules, gene functions, and discover binding sites governing gene expression \citep{Carbonneau:2018}. 
The interpretable attention results lead to detecting meaningful instances (e.g. compounds, molecules, genes) to characterize biological properties. Moreover, the binary classification setting could be easily modified for a multi-class classification setting where the final output is calculated by the softmax function and the loss function is defined by a multi-class cross-entropy for optimization.

The instance selection procedure by the sparse attention could be considered a regularization technique commonly used in statistical learning \citep{hastie2009elements}. The LASSO \citep{tibshirani1996regression} is a representative method to select essential features with sparsity. Regression coefficients of LASSO have non-zero or exact zero values by a shrinkage constraint. The features with non-zero coefficients affect the response variations, but the features with zero coefficients have no influence. The sparsity removes redundant feature information in calculating responses. The proposed method and the LASSO share a similar concept of selecting important information from data. The difference is that LASSO selects important features, but the proposed method selects important instances in a bag. The relationship between the sparse and basic attention for MIL is demonstrated by the relationship between LASSO and Ridge regression \citep{hoerl1970ridge}. Ridge regression is also a shrinkage method, but the coefficients are not forced to be exact zeros. Thus, it is hard to interpret the Ridge regression results regarding feature selection. In the MIL problem, previous methods based on attention are inappropriate for explaining classification results because all the attention values are non-zero.

% We illustrate the attention scores estimated by the softmax and sparsemax functions in Figure \ref{fig:attention}. Moreover, we compare their predictive performance in Section \ref{sec:ablation}.

\section*{Acknowledgement}
\label{sec:acknowledgement}
In loving memory of our late friend Ze Zhang, who was a great company and colleague, we would like to express our deepest gratitude for her kindness and support.

\section*{Funding}
\label{sec:funding}
Younghoon Kim is supported by NRF-2020R1G1A1101853, Tao Wang and Xinlei Wang are funded by NIH (R01CA258584),
Seongoh Park is supported by the Sungshin Women’s University Research Grant of H20200131.

%% The Appendices part is started with the command \appendix;
%% appendix sections are then done as normal sections
% \appendix

% \section{Sample Appendix Section}
% \label{sec:sample:appendix}
% Lorem ipsum dolor sit amet, consectetur adipiscing elit, sed do eiusmod tempor section \ref{sec:sample1} incididunt ut labore et dolore magna aliqua. Ut enim ad minim veniam, quis nostrud exercitation ullamco laboris nisi ut aliquip ex ea commodo consequat. Duis aute irure dolor in reprehenderit in voluptate velit esse cillum dolore eu fugiat nulla pariatur. Excepteur sint occaecat cupidatat non proident, sunt in culpa qui officia deserunt mollit anim id est laborum.

%% If you have bibdatabase file and want bibtex to generate the
%% bibitems, please use
%%
\bibliographystyle{apalike}

\bibliography{auxillary/references.bib}

\newpage
\appendix
\counterwithin{figure}{section}

% \section*{Appendix}
% \section{Extracted features}
% \label{sec:appendix}
% \input{contents/appendix}

\end{document}